\documentclass[]{fairmeta}

\usepackage{listings}
\usepackage{graphicx}
\usepackage{placeins}

\title{EgoToM: Benchmarking Theory of Mind Reasoning from Egocentric Videos}

\author[1,*]{Yuxuan Li}
\author[1]{Vijay Veerabadran}
\author[1]{Michael L. Iuzzolino}
\author[1]{Brett D. Roads}
\author[2]{Asli Celikyilmaz}
\author[1]{Karl Ridgeway}

\affiliation[1]{Reality Labs}
\affiliation[2]{FAIR}

\contribution[*]{Work done during an internship at Meta.}

\abstract{We introduce EgoToM, a new video question-answering benchmark that extends Theory-of-Mind (ToM) evaluation to egocentric domains. 
Using a causal ToM model, we generate multi-choice video QA instances for the Ego4D dataset to benchmark the ability to predict a camera wearer’s goals, beliefs, and next actions.
We study the performance of both humans and state of the art multimodal large language models (MLLMs) on these three interconnected inference problems.
Our evaluation shows that MLLMs achieve close to human-level accuracy on inferring goals from egocentric videos. However, MLLMs (including the largest ones we tested with over 100B parameters) fall short of human performance when inferring the camera wearers' in-the-moment belief states and future actions that are most consistent with the unseen video future.
We believe that our results will shape the future design of an important class of egocentric digital assistants which are equipped with a reasonable model of the user's internal mental states.}

\date{\today}
\correspondence{\email{karl.ridgeway@meta.com}}

\metadata[Code]{\url{https://github.com/facebookresearch/EgoToM}}

\begin{document}

\maketitle

\section{Introduction}
\label{sec:intro}

\begin{figure*}
  \centering
  \includegraphics[width=1.0\linewidth]{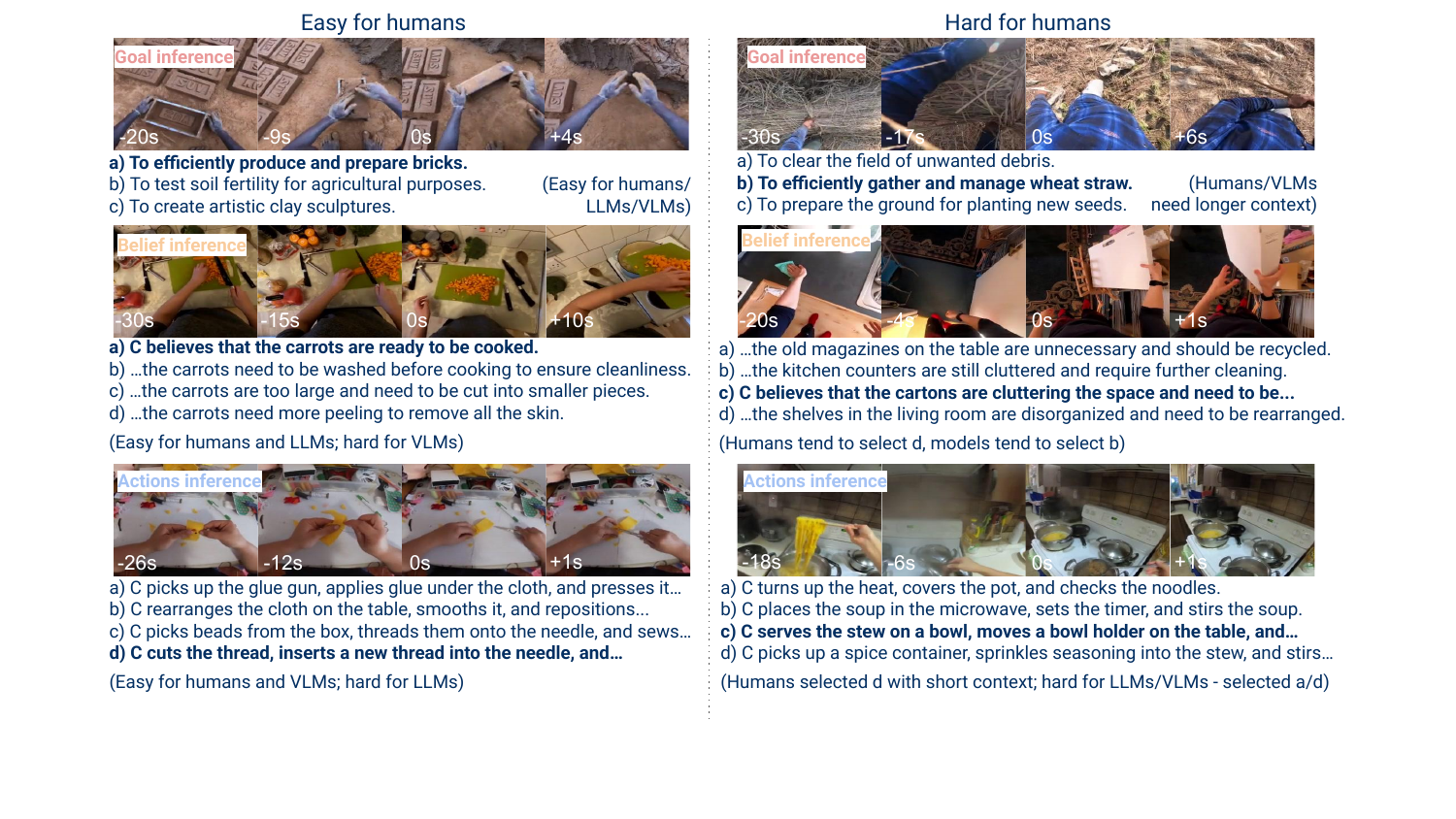}
  \caption{Example questions in \textbf{EgoToM} and difficulty for humans and models, with selected frames before and after the query moment (0s). \textbf{C} indicates the camera wearer. Boldfaced choices are the correct answer best supported by video evidence. For example, in goal inference (right), the single frame at the query moment cannot disambiguate the choices, but the past context suggests that C's main goal is to collect and organize wheat bundles.
  Some statements were slightly edited and trimmed for visualization purposes.
  }
  \label{fig:example}
\end{figure*}

Following the success of large language models, new data, objectives, and alignment methods are bringing multimodal large language models (MLLMs) up to speed on abilities such as generation, captioning, and even visual reasoning and long-context understanding. Among research that aims to test the extent to which these models truly understand visual scenes, a strong emerging interest is to assess whether MLLMs can correctly perform social reasoning based on observed human behavior, such as their ability to reason about a Theory-of-Mind (ToM).

Theory-of-Mind (ToM) refers to the ability to infer unobserved mental states that reflect what others want, think, or believe, and use these states to predict others' future behavior \citep{premack1978does}. In humans, studies have found that the ability to disentangle others' internal beliefs from reality and from one's own beliefs typically develops around the age of 4, with some evidence suggesting an implicit understanding in as early as 15-months of age \citep{onishi200515, wellman2001meta}. Until recently, testing ToM-capability of machine-learning models has largely been relegated to grid-world-type domains where full ground truth of the observed agent is available \citep{baker2009action, rabinowitz2018machine, zhi2020online, gandhi2021baby}, or in restricted domains with a limited set of possible goals and actions, such as web browsing \citep{jiang2021learning}.

The advancements in large-scale foundation models, including language-only models (LLMs) and vision-/video-language models (VLMs), bring the possibility that ToM reasoning may no longer be a uniquely human ability. Recent work has designed story-based benchmarks \cite[e.g.][]{gandhi2024understanding, le2019revisiting} to test whether LLMs can accurately answer ToM-related questions, finding that some LLMs can reach human-level performance in easy conditions but fail in harder conditions. Recent video-based benchmarks have also proposed ways to assess VLMs' analysis of actions beyond simply predicting the correct actions \cite[e.g.][]{du2024towards, kokomind2023, li2023intentqa, mangalam2023egoschema, xiao2021next}. Evaluations of various models have often found that they fall far behind human performance.

Importantly, ToM reasoning is crucial for intelligent egocentric digital assistants to deliver accurate user reasoning and prediction. But a ToM evaluation, especially surrounding the ability to reason about unobserved, internal belief states of the camera wearer, is significantly lacking in egocentric visual domains.  To make up for this gap, we constructed a benchmark to evaluate first-order ToM reasoning of naturalistic human behavior in egocentric videos.

\begin{figure*}
  \centering
  \includegraphics[width=0.95\linewidth]{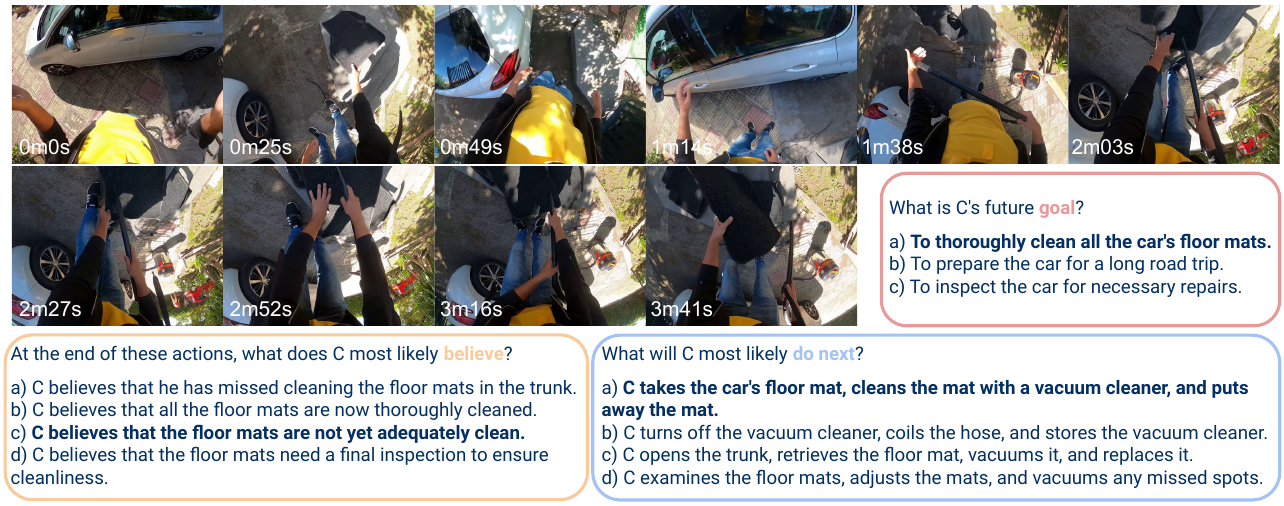}
  \caption{Example questions associated with the same query moment in \textbf{EgoToM}. The questions are queried at 3m41s for this instance.}
  \label{fig:singleq}
\end{figure*}

We introduce \textbf{EgoToM}, an egocentric ToM benchmark that contains a set of multi-choice questions that evaluate a model's coherent understanding of the camera wearer's goals, beliefs, and future actions associated with specific moments in egocentric recordings (see example questions in Figures~\ref{fig:example} and~\ref{fig:singleq}). We propose a structured QA generation pipeline based on a causal ToM behavioral generation model, which allowed us to extrapolate ground-truth goal, belief, and action statements from narrations of atomic actions based on the video future (Figure~\ref{fig:pipeline}). We apply the same causal model to collect realistic counterfactual statements to serve as hard negatives to the ground-truth choices.

EgoToM currently contains over 1k questions (351 goal, 334 belief, and 354 action inference questions) constructed over moments in 785 unique video clips extracted from Ego4D \citep{grauman2022ego4d}. Our evaluation (see Figure~\ref{fig:accuracy}) shows that human participants are adept at first-order ToM inference from egocentric views across all question types (goals, beliefs, and actions). Some LLMs (using human-generated video narrations as the text context) and VLMs (closed- and open-weight models with sample frames as the video context) reach close to human-level goal inference accuracies, but are much less accurate in inferring camera wearers' in-the-moment beliefs and future actions that satisfy the corresponding high-level goals. 

In what's below, we review our benchmark generation pipeline in greater detail, and discuss some important trends of inference accuracies that we observe across models and across multiple conditions that may illuminate future improvements in model ToM inference capabilities. To preview, VLMs achieve higher performance than LLMs in low context conditions, suggesting that action-centric narration is not enough and ToM inference can benefit from rich visual information. Models also do not always benefit from longer context history, suggesting room for better integration over longer contexts. We also find that choice consistency is high between the belief and action inference questions, suggesting that improvement in belief inference can bring improvement in action prediction, and vice versa.

\section{Related Work}

\textbf{Benchmarking Theory-of-Mind.} There has been an increasing interest and availability of benchmarks that target Theory-of-Mind (ToM) reasoning capabilities in large language and multimodal models. For example, ToMi \citep{le2019revisiting} and BigTOM \citep{gandhi2024understanding} are representative LLM benchmarks that both contain ToM reasoning questions based on controlled text story contexts.  Subsequent evaluations generally find that LLMs suffer at zero-shot ToM inference, but their performance can be improved by various reasoning strategies \citep{gandhi2024understanding, sclar2023minding, wilf2023think}.  Recently, new benchmarks have been proposed to target general social reasoning abilities, including efforts such as \citet{kokomind2023} which contains ToM question based on movie contexts.

\textbf{Understanding and predicting human behavior in videos.} General human activity understanding has been a core topic in evaluating model video understanding capabilities. Many benchmarks have been proposed over the years for this purpose, including ActivityNet \citep{caba2015activitynet} and MVBench \citep{li2024mvbench}.  Several benchmarks have also developed questions that target the ability to infer the intentions behind certain goals or actions, e.g., NExT-QA \citep{xiao2021next} and IntentQA \citep{li2023intentqa}.  Other work such as EgoSchema \citep{mangalam2023egoschema} and EventBench \citep{du2024towards} introduced questions that evaluate goal and action-related inference over longer video contexts.

Recent egocentric benchmarks have also proposed new ways to evaluate goal understanding and action prediction from a first-person viewing perspective. For example, the original Ego4D dataset contains future action prediction tasks \citep{grauman2022ego4d}. Ego4D Goal-Step \citep{song2024ego4d} and EgoPlan-Bench \citep{chen2023egoplan} further build on Ego4D to support evaluation of multi-level goal inference and goal-conditioned action prediction. Among existing egocentric benchmarks, EgoTaskQA \citep{jia2022egotaskqa} is perhaps closer in flavor to our work, containing a set of questions that assess model reasoning of actions, intents, goals, and beliefs constructed on a causal trace of actions and object states in a scene. However, questions in EgoTaskQA focus more on atomic action and object states, and can appear less naturalistic due to deliberate obfuscation of object information.  We therefore see \textbf{EgoToM} as complementing these existing benchmarks, focusing on higher-level reasoning and understanding of general task-progress related to egocentric, human goal-directed behavior.
\section{Creating EgoToM}
\label{sec:methods}

\begin{figure*}[t]
  \centering
  \includegraphics[width=1.0\linewidth]{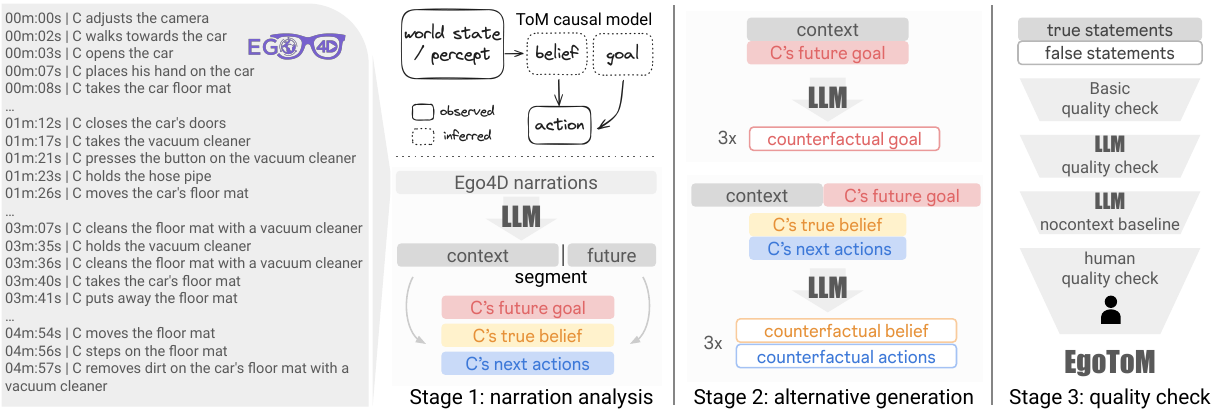}
  \caption{The \textbf{EgoToM} generation pipeline. Given a query moment, ground-truth goal, belief, and actions are extrapolated using future action narrations. Wrong choices are generated only using narrations prior to the query moment and do not align with the video future.}
  \label{fig:pipeline}
\end{figure*}

Unlike story-based Theory-of-Mind (ToM) QA datasets, generating high-quality QA data that assess ToM reasoning but are also visually grounded in realistic videos is challenging, in part due to the absence of accurate text description of unobserved mental states.  Although manual curation of such text descriptions is possible, we draw insights from prior work that makes use of a causal model of the relationship between goal, belief, and action states to generate systematic ToM stories \citep{gandhi2024understanding}. Our key insight is to apply the ToM based causal behavioral template to the dense annotations of real-world human actions available in the Ego4D dataset, combined with an LLM-powered, structured, multi-stage QA generation pipeline \citep{mangalam2023egoschema}. Figure~\ref{fig:pipeline} demonstrates our QA generation pipeline. In what's below, we use C to refer to the camera wearer.

\subsection{Source videos}
We first generated a pool of 5-minute clips from Ego4D videos to build our benchmark on, based on the start/end markers for each \texttt{\#summary} tag (as these were the original 5-min videos supplied to the narrators). We filtered out 7k source clips based on a set of density, diversity, and certainty criteria. Specifically, clips that were least 300 seconds, contain between 60 to 180 lines of narrations without uncertain annotations as marked by \texttt{\#unknown} or \texttt{\#unsure}, and finally with at least 60\% of the narrations about the camera wearer, were selected.  We included both \texttt{narration\_pass\_1} and \texttt{narration\_pass\_2} as narration sources to bootstrap different questions from the same video clip.

\subsection{QA generation}
We made use of LLMs to mass-generate potential choices for three target ToM variables: goal, belief, and actions, following a structured, multi-stage QA generation pipeline. The stages and prompts were developed over multiple iterations.  We primarily used \texttt{GPT-4-Turbo-2024-04-09} (GPT-4T) to generate QA statements.  Full prompts are provided in the Appendix.

\textbf{Narration analysis.}
We first instructed GPT-4T to perform an in-depth narration analysis (Figure~\ref{fig:pipeline}, Stage 1).  We supplied the full list of timestamps and text narrations of each video clip in-context, asking GPT-4T to identify: 1) an interesting moment in the video, and based on this moment, 2) a brief statement of C's future goal, 3) a summary statement of three key actions C takes after this moment, and finally, 4) a statement about C's belief that led C to these particular next actions.  We provide a one-shot example response (without the corresponding narration context).  The statements generated in this narration analysis stage serve as the ground-truth choices in the final choice sets.

\textbf{Counterfactual generation.}
The ground-truth goal, belief, and action statements generated during narration analysis were used to systematically generate counterfactual, or alternative, statements in the next stage to serve as wrong choices in the choice sets (Figure~\ref{fig:pipeline}, Stage 2).  Critically, the generation of the alternative statements also followed the ToM causal template.  For these counterfactual proposals, we now only supply the narration list up until the segmented moment in the context from Stage 1, and instruct GPT-4T to propose realistic alternative goals, beliefs, and actions that C may have that are different from the actual goal, belief, and actions. Although alternative goal generation was more open-ended, we conditioned alternative belief and action statement generation to be conditioned on the true goal statement, to ensure the wrong statement generation also adheres to the causal model and largely satisfy the overall goal. 

In early iterations, we noticed that the alternative action statements generated by GPT-4T consistently included more adverbs than the original human annotations. We thus made an additional round of GPT-4T query to simplify the action statements and improve their style consistency with the ground-truth statement, using a one-shot pair of the original sentence and a simplified version in the context.  The resulting statements closely aligned in style with the Ego4D human narrations.

\subsection{Quality filtering}
\textbf{Keyword-based filtering.}
We noticed that GPT-4T frequently collapsed to a few notable modes in (typically the last) counterfactual goal proposal that resemble the following: To do something for a tutorial/demonstration/class/competition.  We thus removed alternative goal statements that contain these unlikely goals using keyword-based filtering (see the Appendix) and restricted goal inference to three-choice problems.

\textbf{Automatic quality check.}
We used a set of scripted quality checks to remove 1) trials where the segmentation point occurred before 2m30s or after 4m30s, 2) trials where the ground-truth statement is more than 1.4x longer than all three wrong choices or if all three wrong choices were more than 1.4x longer than the ground-truth, and 3) trials where any pair of choices exceed a sentence embedding similarity of 0.94, using \href{https://huggingface.co/sentence-transformers/all-mpnet-base-v2}{\texttt{all-mpnet-base-v2}} \citep{reimers-2019-sentence-bert}.  Together with keyword-based filtering, these automatic quality checks reduced our QA set from 7k clips to 6.4k clips.

\textbf{LLM nocontext baseline.}
We further ensured that the questions involve a visually-grounded context by screening out questions that LLMs can answer correctly by relying on the language prior without any context associated with the video clip.  To capture a wide range of possible language priors that may align with the correct statements without context, we collected responses from \texttt{GPT-4-Turbo} and \texttt{Llama3.1-405b-instruct} for all the QAs without any video or narration-based text context (see prompt in the Appendix).  To ensure that we do not over-exclude trials where the LLMs guessed correct randomly, we collected LLMs' answers for each trial using three randomly-shuffled choice sets, and removed questions where either \texttt{GPT-4-Turbo} and \texttt{Llama3.1-405b-instruct} got more than two times correct out of the three shuffles. This no-context baseline screening reduced our QA set down to 1.7k goal inference questions, 1.2k belief inference questions, and 3.9k action inference questions for further human screening.

\textbf{Human quality check.}
After the automatic quality checks and the LLM nocontext screening, all remaining questions underwent human quality check according to an established rubric (see details in the Appendix). This last round of quality check was performed by the authors, and was meant to ensure that the LLM-proposed choice sets make high-quality and coherent inference problems (e.g., removing QAs where wrong choices capture partially correct goals/beliefs). 
\section{Evaluation}
\label{sec:results}

\begin{figure*}[t]
  \centering
  \includegraphics[width=1.0\linewidth]{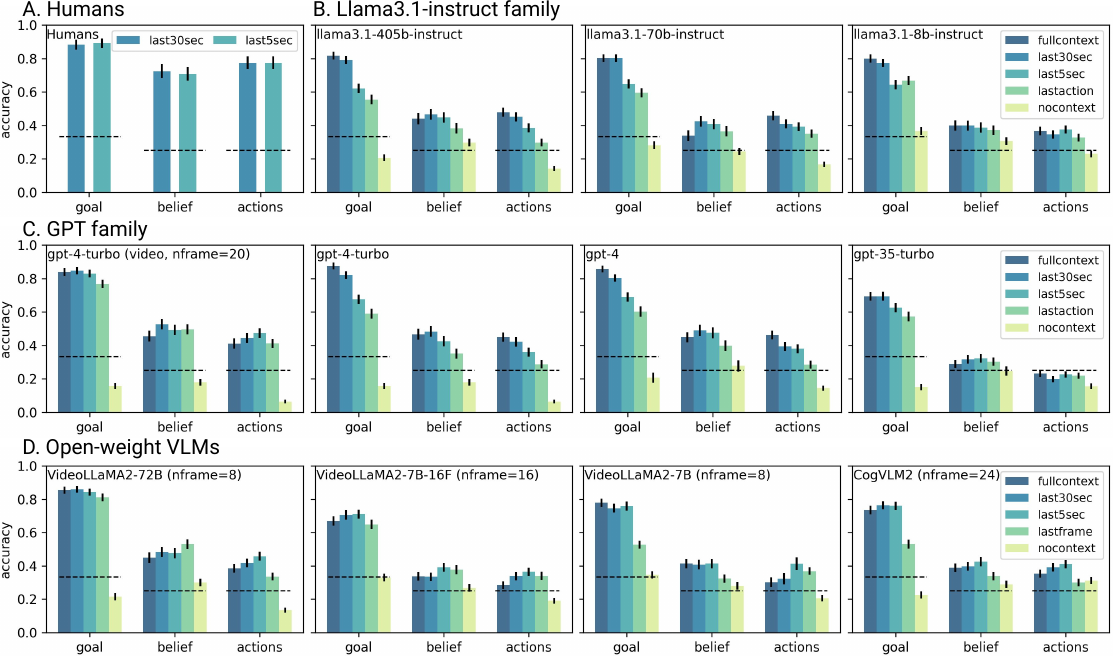}
  \caption{Performance of humans, LLM families, and some open-weight VLMs on \textbf{EgoToM} questions. Dashed lines indicate chance level for each question category. Error bars indicate standard error of the mean (for models, this is computed over the mean accuracy for each question over 3 shuffled choice sets). Legends indicate the context provided for the QAs in each model. Humans, open-weight VLMs, and GPT-4-Turbo are evaluated with sample frames from video contexts. Llama models and GPT models are evaluated with text contexts (Ego4D narrations).}
  \label{fig:accuracy}
\end{figure*}

The multi-choice questions in \textbf{EgoToM} evaluate first-order Theory-of-Mind (ToM) reasoning under realistic, egocentric view of everyday human activities. We evaluate a number of representative state-of-the-art multimodal large language models (MLLMs) and open-weight video-language models (VLMs) in a zero-shot setting. For each question, we evaluate all models on their responses for three randomly-shuffled choice orderings similar to the LLM nocontext baseline (see Section~\ref{sec:methods}).  We supply two types of context: for video contexts, we sample a subset of the video frames as visual context; for text contexts, we use the Ego4D narrations associated with the video contexts.  For video contexts, frame selection is end-aligned (i.e. always includes the last frame) and sampled uniformly (i.e., equidistantly) from the video context, using the standard number of frames for inference for each model type. We vary the amount of context supplied to each question to evaluate how performance changes with more context. The evaluation results presented here are taken from 70\% of the questions in the final benchmark while human quality check was in progress. All models are evaluated with a temperature of 0.

We also collected human performance on a subset of the questions in EgoToM, spanning over 225 unique questions across the three inference categories. Across 4 internal participants who have no prior knowledge of the project, we collected human responses on goal, belief, and action inferences with both the last 5 seconds of video as the context and the last 30 seconds of video as context. Participants watched the videos at 1.5x speed.

Figure~\ref{fig:accuracy} shows our evaluation results. Human responses top the inference accuracy across all three question categories (goal, belief, actions), even when compared to the highest model accuracy. When provided with Ego4D narrations as the text context for the questions, LLMs such as \texttt{Llama3.1-405b-instruct} and \texttt{GPT-4-turbo} show high goal inference accuracy and best belief and action inference accuracies among all models evaluated. When provided with video context, \texttt{GPT-4-turbo} and open-weight VLMs can achieve comparable accuracies to models evaluated with privileged narration information, sometimes even achieving higher accuracies in low-context conditions, suggesting that multimodal models have achieved basic visual processing and language integration that can aid ToM reasoning.

\textbf{Goal inference is easy; belief and action inferences are hard.}
Inferring the camera wearer's future goal is the easiest among the three question types (Figure~\ref{fig:accuracy}). This is consistent across all humans and models we evaluated, where goal inference accuracy can reach more than 80\% for most models (provided there is sufficient context) and humans getting close to 90\% (chance: 33\%). This stands in stark contrast to belief and future action inference, as the models across all text and video context conditions only achieve about 50+\% at best (chance: 25\%). Humans, however, can effectively infer the correct belief and next actions for 70\% and 77\% of the questions.

The difficulty of belief and action inferences likely reflects their level of specificity and tighter association with the ending moment in the given context.  Goal inference concerns more higher-level inference that is less strictly time-locked to a particular moment, and may be more easily discerned conditioned on broad information in the context (whether through text or video frames). However, the wrong belief and action statements are generated conditioned on the true goal, and represent likely counterfactual futures that could have happened but didn't in the particular video instances. Thus, inferring the correct belief and next actions requires a level of sensitivity of the detailed behavior over time that accumulates signals for the true future behavior.  This makes the belief and action inferences hard problems that even humans do not reach ceiling performance, although humans show a large leading gap for both question types compared to the best model performance.  We further reflect on the difficulty of belief and action inference in the Discussion section.

\textbf{The context effect $\times$ context modality.}
As we increased the context shown for answering all question types, models generally show improved accuracies (Figure~\ref{fig:accuracy}). This context effect is most prominent when the context is provided through text, as seen in \texttt{GPT} and \texttt{Llama3.1} family models, although it does not always lead to monotonically increasing belief and action inference accuracies (e.g., beyond 30 seconds prior to the query moment).  The context effect is a lot weaker for models receiving video context.  We note that this is in part due to models receiving the same number of equidistant frames (also referred to as uniform frame sampling) across video contexts of different lengths. As a result, rather than receiving \textit{more} video context, VLMs receive the same amount of information sampled more \textit{sparsely} from a longer video context. 

Even though the increasing context effect is stronger for text contexts, models achieve noticeably higher accuracies with a few video frames in low-context conditions than a few lines of human narration texts (i.e., compare the lastframe condition with the lastaction condition in Figure~\ref{fig:accuracy}). For example, LLMs struggle in Figure~\ref{fig:example}'s left action inference problem due to lack of narration of the final moment in the video context showing that C reaches for the scissors. This suggests that multimodal models can effectively pick up useful information beyond simple narrations, and reflects that human narrations are an inherently lossy compression of the rich visual information in a video. Thus, the weaker context effect may stem from a weakness in integrating visual information across multiple frames or over time. This is seen, for example, when comparing \texttt{VideoLLaMA2-7B} with \texttt{VideoLLaMA2-7B-16F}, where sampling 16 frames as the visual context did not show improvement over 8 frames (see Figure~\ref{fig:accuracy} bottom row).

To our surprise, average human responses showed virtually no gain in extending from a 5-sec context to a 30-sec context. One possibility is that humans are highly efficient at ToM-related reasoning, although this may be in part due to the small sample size of participants and only using a subset of questions. Upon closer inspection, we observed that humans were able to select the correct answer for some questions with lengthened context sometimes, but also wrongly selected away from the correct answer they chose under the shorter video context. Assimilating information from longer contexts may have distracted participants from the fact that the belief and action inferences are specifically tied to the ending moments of the video context, and presents a general challenge for being simultaneously able to understand long-context information and accurately infer in-the-moment mental states and next actions of the camera wearer.

\textbf{(Language) Model scaling generally helps.}
Comparing model variants within a model family, we also noticed that scaling model sizes generally improves accuracy across goal, belief, and actions inferences. For example, \texttt{Llama3.1-405b-instruct} is better than \texttt{Llama3.1-70b-instruct} and \texttt{Llama3.1-8b-instruct} at belief and action inferences, and \texttt{GPT-4-turbo} and \texttt{GPT-4} are noticeably better than \texttt{GPT-35-turbo} across all question types, as \texttt{GPT-35-turbo} only achieves chance-level belief and action inference even under large amount of text context. For open-weight VLMs, for example, \texttt{VideoLLaMA2-72B} achieves overall higher accuracies in all three question types than the \texttt{VideoLLaMA2-7B} variants. We note however that variants within a single model family can sometimes differ in many significant aspects than simply the number of parameters, and for that reason these results only loosely capture a true scaling effect. For video-language models, it would also be interesting to test the scaling effect for model variants that differ in the visual encoder size but have same-sized LLM components, which we hope to pursue as future work.

\begin{figure*}[!htb]
  \centering
  \includegraphics[width=1.0\linewidth]{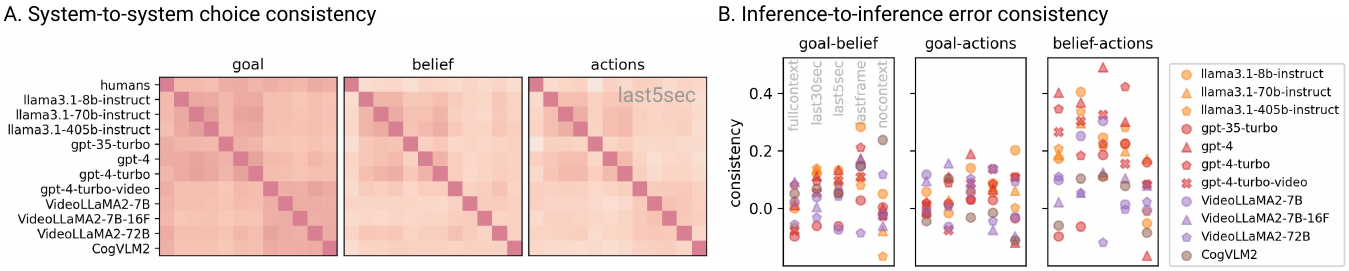}
  \caption{A. Choice consistency between humans and models. B. Error consistency between inference types within each model. Error consistency measures binary answer correctness alignment between two response profiles.  Choice consistency measures multi-way choice selection alignment between two response profiles.}
  \label{fig:consistency}
\end{figure*}

\textbf{Choice consistency across humans and models.}
We evaluated whether selected choices in different questions aligned across humans and models, using a generalization of the error consistency metric referenced in \citet{geirhos2020beyond} that accounts for alignment due to chance. 
Specifically, the consistency metric $\kappa$ between two response profiles $p$ and $q$ across a set of shared questions is given as:
\begin{align}
    \kappa_{p,q} = \frac{c_{obs_{p,q}} - c_{exp_{p,q}}}{1 - c_{exp_{p,q}}}
\end{align}
where $c_{obs_{p,q}}$ is the mean observed matching responses across $n$ samples:
\begin{align}
    c_{obs_{p,q}} = \frac{1}{n} \sum_{i=1}^{n} \mathbf{1}(p_i=q_i)
\end{align}
and $c_{exp_{p,q}}$ measures matched responses due to chance for $m$ unique labels:
\begin{align}
    c_{exp_{p,q}} = \sum_{j=1}^{m} P(p_i=j) P(q_i=j)
\end{align}
For multi-class choice consistency, the response profiles are the selected choice indices. For error consistency, the response profiles are the binary correct vs. incorrect score of the selected choices.

Figure~\ref{fig:consistency} shows the human-to-model (using aggregated responses from all humans) and model-to-model choice consistency, along with inference-to-inference error consistency within the same model across inference categories. Figure~\ref{fig:example} also contains notes on choice consistency for some example trials.

When comparing model choices to human choices, we observed the highest choice consistency for inferring camera wearers' future goals, and lower consistency for belief and action inference (Figure~\ref{fig:consistency}A). This in part reflects the higher goal accuracies in both humans and models and larger performance gap in belief and action inferences. Across different models, we observed high consistency for goal inference compared to belief and action inference, and this consistency is sometimes clustered by the type of contexts (text or video, see Figure~\ref{fig:consistency}A).

Some video clips and their query moments are associated with multiple question types, allowing us to compare error consistency across different inference problems on the same context within each model (Figure~\ref{fig:consistency}B). The results generally show a higher consistency between belief and action inference for the same text or video context. This is consistent with the fact that both the ground-truth and alternative belief and action statements are generated, by design, in a paired fashion.
\section{Discussion}
\label{sec:discussion}

We present \textbf{EgoToM}, a benchmark designed to evaluate first-order Theory-of-Mind (ToM) reasoning of naturalistic human goal-directed behavior as observed in egocentric videos. The paired goal, belief, and action questions make EgoToM a unique opportunity to evaluate accurate and consistent ToM reasoning at multiple stages in the causal behavioral model. Our evaluation results suggest that large language models and multimodal language models do reasonably well on inferring human future goals, but are yet to excel at inferring camera wearers' in-the-moment internal belief states and predicting camera wearers' future actions. 

Our key insight is to apply a ToM causal model as a behavioral generation template to interpreting human behavior as described in the human narrations in Ego4D videos. This allows our benchmark generation to 1) leverage a structured interpretation of sequential behavior to extrapolate the unobserved and un-annotated internal belief states of the camera wearer, and 2) generate counterfactual, paired belief and action statements conditioned on the same true goal to use as hard wrong choices in the multi-choice QA set.  We propose that this general framework can be useful in much broader settings to scale data collection beyond human annotations. However, such generation can certainly lead to non-visually-grounded text data. In building the benchmark, we implemented strict human screening to ensure high-quality, visually-grounded choice statements. The problem of visual grounding will likely be remedied as the field improves the abilities of language models and multimodal models. To this end, using a structured pipeline corresponding to a causal model of behavior generation may serve as an initial part of the flywheel, where the true and false statements collected in early rounds can be used to aid model pre-training and fine-tuning.

In EgoToM, belief and action inferences are hard as the wrong choices were constructed to represent realistic alternative futures that do not align with the ground-truth video future and are less-well supported by video evidence. Importantly, human mental states are nuanced and multi-level, thus camera wearers in the videos can reasonably have multiple beliefs. During benchmark generation, our human quality check stage strove to remove questions where wrong choices capture partially correct goals or beliefs. But a true vs. false distinction based on an idiosyncratic behavioral trace is nonetheless sometimes rigid. We thus see two ways to use the benchmark: 1) to evaluate against the correct choice based on the video ground-truth future, and 2) to assess alignment with a human choice distribution. We see the latter approach on evaluating model-human choice alignment as a promising future direction to explore.

Although some questions in EgoToM can be approximately solved using a few last frames, often times discerning between the alternative choices requires integrating snippets of behavior further back in the context relative to the moment being queried (e.g., to observe that the camera wearer has already checked the oil tank in a lawn mower earlier and thus would not have the belief that the oil level is insufficient).  As expected, we observed that including more visual or narration context increases inference accuracy.  However, as we briefly discussed above, one difficulty in evaluating model ability to infer goals, belief, or actions under a video context is the issue of frame selection.  Indeed, selecting a critical subset of the frames is a part of the inference problem. Here, we sampled a number of equidistant frames consistent with the inference procedure provided in the model cards or default inference code.  Although this method represents the standard inference mode of these models, this may not reflect the full capabilities of these models to reason over long video contexts. 

We also note a few idiosyncratic features of the belief and action inference problems that may have contributed to the performance seen in humans and models.  First, due to the inherent short time-scale of each human narration entry, different objects across frames are sometimes referred to using the same phrase (e.g., “the dough” when there are multiple doughs). This can occasionally lead the object being referenced in different choices ambiguous.  Second, we relied on the human narrators' timestamps to crop the video context for the problems. Due to the inherent inaccuracy and difficulty in aligning the exact timing of the start of an action when human annotators annotated Ego4D videos, sometimes the initiation of the true next action is observable in the last few frames of the video context. Humans may have more effectively leveraged this “leaking action information” in low-context conditions than VLMs, contributing to the action inference gap seen across the two.

EgoToM is mostly a first-order ToM benchmark, in that models reason about the goals, beliefs, and actions directly associated with the camera wearer. However, some trials in EgoToM do capture social interactions between the camera wearer and other humans, and as a result these QAs implicitly measure an ability to reason about how the camera wearer perceives and interprets others' mental states.  In future work, it would be interesting to apply this benchmark generation pipeline to more complex, higher-order ToM inference in videos of richer social interactions. Additionally, it might be concerning that humans do not achieve perfect accuracy on goal, belief, or action inference, but we should not expect them to: humans are not perfect observers. For example, human performance on story-based ToM belief inference is only at 70\% with a 50\% chance rate \citep{gandhi2024understanding}, while EgoToM belief inference is at 70\% with a 25\% chance rate.
It is also worth noting that human accuracy at inferring actions is higher than inferring beliefs in our benchmark. As noted above, some final video frames may leak initiation of the actual future actions. It may be also possible that humans do not always elicit an expensive ToM reasoning for inferring other humans' future behavior.

Ultimately, we view this work as taking a first step in helping us think about how to build the next generation of context-aware egocentric digital assistants that not only predict future goals and actions, but do so by reasonably tracking the users' internal mental states. Looking ahead, we envision research work in both examining the extent to which ToM inference can come about from pure action prediction across diverse behavioral trajectories, and effort towards more direct training to improve reasoning abilities related to unobserved mental states. We hope that the present work provides useful evaluation and inspiration in both cases.

\bibliographystyle{assets/plainnat}
\bibliography{main}

\clearpage
\newpage

\beginappendix
\section{All prompts used}

\subsection{Benchmark generation}

\subsubsection{Narration Analysis Prompt}
\begin{lstlisting}
You are an expert at analyzing human behavior. Answer the questions based on the template. Provide short, assertive, single-sentence responses.

Analyze the video narrations of a human actor, C. Answer the following questions.
1. What is an interesting transition point where C moves on to the next set of actions? The transition point must be in the latter half of the narrations. Provide the exact timestamp and the narration text.
2. What is C's future goal across all actions after the transition point? The goal statement must be high-level and brief.
3. Summarize three key actions C takes at and after the transition point into a coherent action plan. Only include actions essential to C's goal.
4. What is C's belief at the transition point that led C to these next actions? The belief statement should focus on C's belief about object states or task progress, without mentioning specific actions.

Example response:
Transition point: 03m:18s | #C C picks a pair of scissors.
Future goal: To sew a piece of cloth.
Next actions: C picks scissors, trims the thread, and passes the thread through the needle.
C's belief: C believes that the thread is too thick for the needle's eye.

Provide your responses by analyzing the following narrations:
Video summary: {clip_summary}
{narrations}

Response:
\end{lstlisting}

\subsubsection{Counterfactual goal generation prompt}
\begin{lstlisting}
You are an expert at predicting human behavior and generating counterfactual scenarios. Answer the questions based on the template. Your responses should be creative but plausible.

Analyze the video narrations of a human actor C doing household, outdoor, workplace, or leisure activities. Propose three wrong goals at the end of theses actions that are different from C's actual goal. The correct and wrong goal statements will be used as choices in a multi-choice test. Make it difficult to find the correct statement among all the choices.
The goal statements must be high-level and brief. Few words are enough.

Narrations of C's actions: 
{narrations_in_context}

Correct goal: {gt_goal}
Wrong goal 1:
Wrong goal 2:
Wrong goal 3:
\end{lstlisting}

\subsubsection{Counterfactual belief and action generation prompt}
\begin{lstlisting}
You are an expert at predicting human behavior and generating counterfactual scenarios. Answer the questions based on the template. Your responses should be creative but plausible.

Analyze the video narrations of a human actor, C. Given C's goal and C's actual belief, propose different beliefs C may have at the end of the video narrations and the appropriate actions C should take under each belief. These alternative belief and action statements will be used as choices in a multi-choice test. Make it difficult to find the actual statement among all the choices.
The belief statements should focus on object states or task progress, without mentioning specific actions. The action statements should describe three simple, physical actions in one sentence, focusing on movement and interaction with objects.

C's goal: {gt_goal}
Narrations of C's actions: 
{narrations_in_context}

Actual belief: {gt_belief}
Actual next actions: {gt_actions}

Alternative belief 1:
Alternative next actions 1:

Alternative belief 2:
Alternative next actions 2:

Alternative belief 3:
Alternative next actions 3:
\end{lstlisting}

\subsubsection{Action statement simplification prompt}
\begin{lstlisting}
You are an expert at extracting core meanings in sentences.

Simplify each of the following sentences to be more concise. Remove details like adverbs or action purposes that do not change the meaning of the sentences. The resulting statements should describe three simple physical actions, focusing on movement and interaction with objects.

Example sentence: C adds more water to the mixture, stirs thoroughly to achieve the right consistency, and tests the mortar by applying a small amount to a brick.
Example simplified sentence: C adds water to the mixture, stirs mixture, and tests the mortar on a brick.

Sentence 1: {alt_actions_1}
Sentence 2: {alt_actions_2}
Sentence 3: {alt_actions_3}

Simplified sentence 1:
Simplified sentence 2:
Simplified sentence 3:
\end{lstlisting}

\subsection{Evaluation}
\subsubsection{Nocontext Prompt}
\label{prompt:nocontext}
\begin{lstlisting}
You are an expert at predicting human behavior. Select the best answer for each question. Answer with your best guess even when there is not enough information. Answer as 'Answer 1: <option>) <answer>'.

C is a person.
Question 1: What does C most likely believe?
  a) {belief_choice_a}
  b) {belief_choice_b}
  c) {belief_choice_c}
  d) {belief_choice_d}
\end{lstlisting}

\section{Keyword-based filtering for alternative goal statements}
We used the following keyword list: \texttt{['demonstrate', 'demonstration', 'tutorial', 'class', 'train for', 'compete', 'competition', 'technique']}.

\section{Human quality check rubric}
\label{humanqualitycheckrubric}
\begin{lstlisting}
- Video quality is good. Skip all following for a bad video.

- Watch the video until the segmented transition timepoint and a bit afterwards. You can drag the progress bar, but make sure to understand the video context well. You may need to rewatch the video after seeing the GT + alternative statements for more details.

- The transition point has potentially interesting inferences. The transition point must be relevant to some minimal amount of shift in behavior, e.g., if the entire clip is continuing driving/walking or pruning a bush, there is little room for inference. However, the transition point doesn't always need to be "move on to something else". Some continuing actions may be interesting, if the ground-truth belief is about "task is incomplete" (e.g. the food needs more cooking) and the alternatives include "task is complete" (e.g. the food is ready).

- The ground-truth statements are accurate and contextualized. 

- The ground-truth statements should be inferrable from the video context up until the transition point.
    - It is okay if the goal is brief and does not differentiate the goal going-forward vs. the goal for the entire clip.
    - gt_belief should accurately capture C's intent and is coherent with gt_goal and gt_actions.
    - gt_belief should be tied to the transition point, not to an earlier or later moment in the video.
    - gt_actions should be relevant to the goal/belief.

- Alternative statements make good multi-choice problems. 
    - The statements should be sufficiently different from each other. If some alt statements are lower-quality, e.g. too broad or obviously untrue, that's okay.
    - Sometimes if the GT isn't all that accurate, a question can still be good if the GT statement is still the best description because the alternatives are more wrong
    - Many alternative goals are of the type "to do X for a class/tutorial/demonstration" "to train for a competition". For now, count these as good alternatives. We'll do a round of keyword-based filtering to remove those alternative goals and make goal inference three-choice problems.
    - If the ground-truth + alternatives make a good set of multiple choices, mark the corresponding columns [goal_question, belief_question, action_question] as "g" (good). Otherwise, mark "b" (bad), or "u" for undecided.
\end{lstlisting}

\end{document}